\documentclass[10pt,twocolumn,letterpaper]{article}

\usepackage{iccv}
\usepackage{times}
\usepackage{epsfig}
\usepackage{graphicx}
\usepackage{amsmath}
\usepackage{amssymb}

\usepackage{amsmath,bm}
\usepackage{amsfonts,amssymb}
\usepackage{booktabs}
\usepackage{graphicx} 
\usepackage{float}
\usepackage{subfigure}
\usepackage{bm}
\usepackage{color}
\usepackage{multirow}

\usepackage{widetext}

\newcommand{\vts}{\mathrm{T}}
\newcommand{\bv}[1]{{\bm {#1}}}
\newcommand{\ba}[1]{{\bf {#1}}}
\newcommand{\beq}{\begin{equation}}
\newcommand{\eeq}{\end{equation}}
\newcommand{\bmat}{\begin{bmatrix}}
	\newcommand{\emat}{\end{bmatrix}}
\newcommand{\Eq}[1]{(\ref{#1})}
\newcommand{\Fig}[1]{Fig. \ref{#1}}
\newcommand{\Tab}[1]{Tab. \ref{#1}}

\usepackage{booktabs}
\usepackage{caption}
\usepackage[T1]{fontenc}
\usepackage{wrapfig}
\usepackage{multirow}
\usepackage{varwidth}
\usepackage{caption}
\usepackage{colortbl}
\usepackage{xcolor}
\usepackage{tabularx}
\usepackage{xspace}
\makeatletter
\DeclareRobustCommand\onedot{\futurelet\@let@token\@onedot}
\def\@onedot{\ifx\@let@token.\else.\null\fi\xspace}

\makeatother
\usepackage[breaklinks=true,bookmarks=false]{hyperref}

\iccvfinalcopy % *** Uncomment this line for the final submission

\def\iccvPaperID{10268} % *** Enter the ICCV Paper ID here

\begin{document}
	
% \hypersetup{hidelinks}

%%%%%%%%% PAPER BEGIN
%%%%%%%%% TITLE
\title{StructDepth: Leveraging the structural regularities 
	for self-supervised indoor depth estimation}

%% our
\author{Boying Li$^{*}$, Yuan Huang\thanks{Both are the first authors with equal contributions. $^{\dag}$Corresponding author: Danping Zou ({\tt\small dpzou@sjtu.edu.cn)}. This work was supported by NSFC (62073214).} , Zeyu Liu, Danping Zou$^{\dag}$, and Wenxian Yu\\
	\emph{Shanghai Key Laboratory of Navigation and Location-Based Services}\\
	\emph{Shanghai Key Laboratory of Intelligent Sensing and Recognition}\\ %, Shanghai Jiao Tong University\\
	Shanghai Jiao Tong University\\
} 

\maketitle
% Remove page # from the first page of camera-ready.
%%% \ificcvfinal\thispagestyle{empty}\fi

%%%%%%%%% ABSTRACT
\begin{abstract}
	Self-supervised monocular depth estimation has achieved impressive performance on outdoor datasets. Its performance however degrades notably in indoor environments because of the lack of textures. Without rich textures, the photometric consistency is too weak to train a good depth network. Inspired by the early works on indoor modeling, we leverage the structural regularities exhibited in indoor scenes, to train a better depth network. Specifically, we adopt two extra supervisory signals for self-supervised training: 1) the Manhattan normal constraint and 2) the co-planar constraint. The Manhattan normal constraint enforces the major surfaces (the floor, ceiling, and walls) to be aligned with dominant directions. The co-planar constraint states that the 3D points be well fitted by a plane if they are located within the same planar region. To generate the supervisory signals, we adopt two components to classify the major surface normal into dominant directions and detect the planar regions on the fly during training. As the predicted depth becomes more accurate after more training epochs, the supervisory signals also improve and in turn feedback to obtain a better depth model. Through extensive experiments on indoor benchmark datasets, the results show that our network outperforms the state-of-the-art methods. The source code is available at  \url{https://github.com/SJTU-ViSYS/StructDepth}.  %   Through extensive experiments on indoor benchmark datasets, the results show that our self-supervised network outperforms the state-of-the-art self-supervised ones.
\end{abstract}

%%%%%%%%% BODY TEXT
% fig 1
\section{Introduction}
\begin{figure}[ht]		
	\centering  
	\includegraphics[width=0.43\textwidth]{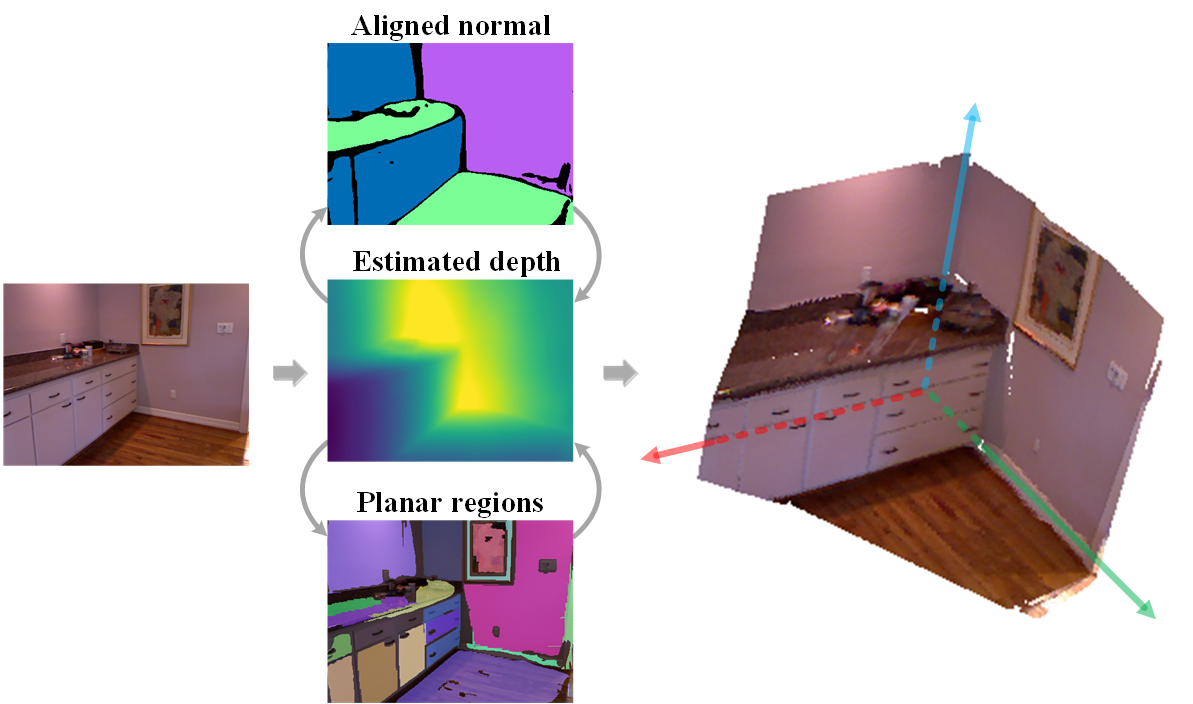} 	% 0.44
	\caption{Our self-supervised monocular depth learning leverages the structural regularities of indoor environments for training. The aligned normal (with Manhattan directions) and the planar regions provide extra losses in training and lead to better 3D structures at inference.}
	\label{fig:main}  
	% \vspace{-0.3cm} 
\end{figure}

Inferring the dense 3D map from a single image is a challenging problem without satisfactory solutions until the booming of deep neural networks. With the deep convolutional neural networks (CNNs), we can predict the accurate depth from a single image, via training the network with a lot of ground-truth depth labels. The recent self-supervised learning paradigm  does not require the ground-truth depth, while still obtaining high-quality results on benchmark datasets, using the photometric consistency as the major supervisory signal. Nevertheless, when existing self-supervised methods are trained on indoor images, the quality of depth estimation degrades notably\cite{zhou2019moving}\cite{bian2020unsupervised}. The main reason is the lack of textures in indoor images. Unlike outdoor scenes, the indoor scenes are full of texture-less regions, such as white walls, ceilings, and floors. Without rich textures, the photometric loss becomes too weak to train a good depth model. Seeking stronger or extra supervisory signals is therefore necessary for training a better depth network.

There have been a few attempts. An optical-flow field propagated from the sparse SURF\cite{bay2006surf} flow by a self-supervised network, is used to guide training on texture-less regions \cite{zhou2019moving}. Another attempt \cite{yu2020p} is to use an image patch instead of individual pixels to compute the photometric loss and apply extra constraints to the depth within the planar regions extracted from image segmentation. Though those attempts improve the results, they did not fully exploit the structural regularities presented in indoor environments, a valuable source of information for 3D learning. The structural regularities, known as the Manhattan-world model\cite{coughlan1999manhattan}, describe that the scene consists of major planes aligned with dominant directions. This simple yet effective high-level prior leads to a much better performance in many vision tasks, such as indoor modeling\cite{furukawa2009manhattan}\cite{furukawa2009reconstructing}\cite{concha2014manhattan}, visual SLAM\cite{zhou2015structslam}\cite{flint2010growing}\cite{yang2016pop}, and visual-inertial odometry\cite{zou2019structvio}, but has not been applied to monocular depth learning.

In this work, we propose to apply the high-level prior of indoor structural regularities to self-supervised depth estimation as shown in \Fig{fig:main}. Specifically, we adopt two extra supervisory signals for training: 1) the Manhattan normal constraint and 2) the co-planar constraint. The Manhattan normal constraint enforces the major surfaces (the floor, ceiling, and walls) to be aligned with dominant directions. The co-planar constraint states that the 3D points be well fitted by a plane if they are located within the same planar region.	We add two extra components into the training process. The first one is Manhattan normal detection. It classifies the major surface normal, computed from the depth predicted by the network, into the directions associated with the vanishing points by an adaptive thresholding scheme. The second one is planar region detection. We fuse the color and the geometric information derived from the depth and apply a classic segmentation algorithm to extract planar regions. 
During training, the two components incorporate the estimated depth to produce supervisory signals on the fly. Though those signals may be noisy in early epochs because of inaccurate depth, they will gradually improve as the depth quality improves, and in turn benefit the depth estimation.

We conduct experiments on the indoor benchmark datasets: NYU-v2 \cite{silberman2012indoor}, ScanNet\cite{dai2017scannet}, and InteriorNet\cite{li2018interiornet}. The results show that our method outperforms the existing state-of-the-art methods. Our main contributions are as follows: 

1) A novel learning pipeline for self-supervised depth estimation leveraging structural regularities of indoor environments. To our best knowledge, this has not been presented in previous work.

2) Two novel components providing extra supervisory signals on the fly during the training process. Our components can be used to train a multi-task network including depth estimation, normal estimation, and planar region detection in a self-supervised manner, although the latter two tasks serve to train a better depth model in our current implementation. 

3) We set a new state-of-the-art in self-supervised indoor depth estimation.

\begin{figure*}[ht]		
	\centering  
	\includegraphics[width=0.92\textwidth]{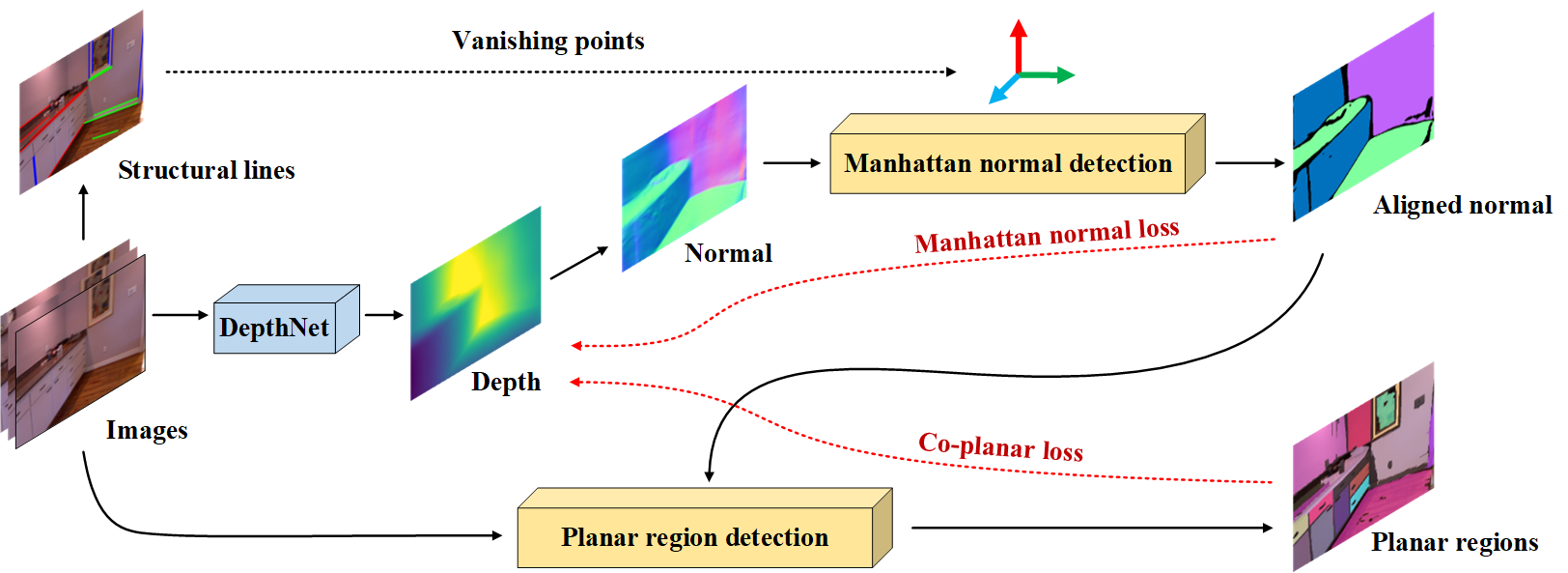} 	% 0.85   0.92
	\caption{Our self-supervised monocular depth learning pipeline, 
		which consists of three major components: 
		a) \textbf{DepthNet}: The neural network to be trained to predict the depth from a single image. 
		b) \textbf{Manhattan normal detection}: It classifies the surface normal estimated from depth prediction into dominant directions.
		c) \textbf{Planar region detection}: Both the color and geometric information are used to extract planar regions by a graph-based segmentation.
		The planar region detection is kept updated with the improved depth during training iterations.
		Two extra losses, \emph{Manhattan normal loss} and \emph{co-planar loss}, are used to train the network, as indicated by the red dot arrows.}
	\label{fig:flowchart}  
\end{figure*}

\section{Related Work}

\paragraph{Monocular depth estimation.} 
Depth estimation from a single image is an ill-posed problem that is known as extremely hard to be solved. Since the pioneer works\cite{eigen2014depth,eigen2015predicting} employed the convolution neural networks (CNNs) to regress the depth directly, a lot of CNN-based monocular depth estimation methods have been proposed \cite{liu2015learning,laina2016deeper,kim2016unified,wofk2019fastdepth,fu2018deep}, producing impressively accurate results in benchmark datasets. Most of them are supervised methods that require the ground-truth depth data for training.

Self-supervised depth learning without the ground-truth depth has emerged as a promising alternative as acquiring the ground-truth depth at a large scale is challenging. The image appearance was firstly introduced in \cite{godard2017unsupervised} to replace the ground-truth depth as the supervisory signal to train a depth network. One image in a stereo pair was warped to the other view by the predicted depth. The difference between the synthesized image and the real image, or the photometric error, is then used for supervision. The idea was further extended to monocular settings \cite{zhou2017unsupervised}\cite{godard2017unsupervised}. By the careful design of network architectures\cite{godard2019digging}, loss functions \cite{shu2020feature}, and online refinement \cite{chen2019self}, self-supervised approaches obtain impressive results on benchmark datasets.

Despite achieving impressive performance on outdoor datasets, such as KITTI\cite{geiger2012we} and Make3D\cite{saxena2008make3d}, existing self-supervised methods perform poorly in indoor datasets. The reason is that the indoor scenes are full of texture-less regions, such as white walls and ceilings, making the photometric loss become too weak to supervise the depth learning.
Zhou et al.\cite{zhou2019moving} adopted an optical-flow-based training paradigm supervised by the flow field from an optical flow network, initialized from sparse SURF \cite{bay2006surf} correspondences. The recent work \cite{yu2020p} employed the more discriminative patches instead of individual pixels to compute the photometric loss, and also applied the piece-wise planar prior to depth learning by assuming that the homogeneous-color regions are planar regions.
Though their approaches improve the performance. They did not fully exploit the structural prior of the environments. In addition, the planar-region assumption in \cite{yu2020p} does not hold for planes with the same color, e.g. mutually perpendicular white walls. It therefore leads to false planar regions deteriorating the depth model.

\paragraph{Planar region detection.}
Though powerful planar-region detectors \cite{liu2019planercnn}\cite{yang2018every}\cite{yu2019single} have been proposed recently and have shown high-quality results in complex indoor images. Those CNN-based detectors require a huge number of plane labels for training and are not suited for the self-supervised learning scheme. Though detecting planes in the image is challenging, if the depth is available, this task becomes much easier\cite{salas2014dense}\cite{kim2018linear}. 
Here, we detect the planar regions using a classic graph-based segmentation approach \cite{felzenszwalb2004efficient} similar to \cite{yu2020p}, while employing the additional geometric information extracted from the depth estimated on the fly when training. Though the depth may not be precise initially, it will gradually improve as the training progresses such that the segmentation will improve as well.
With the additional geometric information, our approach avoids false planar regions that are indistinguishable by colors and produces less over-segmentation on texture-rich planar regions.

%\vspace{-0.1cm} 
\paragraph{Structural regularities in indoor environments.}
Indoor scenes exhibit strong structural regularities, which can be described as the “Manhattan world”. Namely, the scene can be decomposed into major planes, where their normal vectors are mutually orthogonal. These structural regularities are valuable priors that have been applied to a wide range of indoor 3D vision tasks, such as vSLAM\cite{zhou2015structslam}\cite{flint2010growing}\cite{yang2016pop}, VIO\cite{zou2019structvio}, and mapping\cite{furukawa2009manhattan}\cite{furukawa2009reconstructing}\cite{concha2014manhattan}.
In fact, exploiting the structural prior of indoor scenes was probably the only geometric way to infer the 3D information from a single image in early days \cite{delage2006dynamic}\cite{lee2009geometric}. It is natural to think that structural regularities should also benefit the learning-based vision tasks in indoor environments.

%\vspace{-0.1cm} 
Wang et al. \cite{wang2020vplnet} propose to use the vanishing points and lines to train a surface normal estimator which achieves the state-of-the-art performance. Our work adopts a similar spirit but differs from theirs in that our major task is depth estimation, where the surface normal is just an intermediate result that serves for better training. In addition, our depth network is trained in a fully self-supervised manner and does not require the line map as the extra input.
To our best knowledge, our work is the first one incorporating the structural regularities of indoor environments into self-supervised monocular depth estimation.

\section{Method}

Our self-supervised depth learning pipeline is illustrated in \Fig{fig:flowchart}. It consists of three major components. The first one is the depth network, which takes a single image as the input and predicts a depth map. We use the same architecture as in \cite{yu2020p} for the depth network. Based on the predicted depth, the other two components, Manhattan normal detection and planar region detection, are used to produce the supervisory signals leveraging the structural prior of indoor environments.	
Manhattan normal detection aligns the normal computed from the depth map with the dominant orientations, estimated from the vanishing points in the image. Planar region detection applies a graph-based segmentation to detect the planar regions with the combination of color, normal, and plane-to-origin distance information. Both Manhattan normal detection and planar region detection may be inaccurate in the initial training epochs, but they will improve in later epochs as the depth prediction becomes better. The improved supervisory signals lead to a better depth prediction as well.

In the following sections, we'll describe how we apply the Manhattan normal constraint and the co-planar constraint in our training process.

\subsection{Manhattan normal constraint}

\textbf{Dominant direction extraction.}
The structural regularities of indoor environments imply that most indoor scenes contain planar surfaces aligned with dominant directions. The dominant directions can be estimated from the structural lines in the image. The intersection of a set of parallel structural lines in the image is the vanishing point. Let $\bv{v}$ be a vanishing point extracted from the 2D image. One of the dominant directions in the camera coordinate system is computed as
\beq
\bv{\eta} \propto \ba{K}^{-1}\bv{v},
\eeq 
where $\bv{\eta} \in \mathbb{R}^3$ is a unit vector representing this dominant direction and $\ba{K}$ is the camera intrinsic matrix. Note that we need only two vanishing points to get all the dominant directions, since the third dominant direction can be obtained by the cross product. We apply the 2-Line searching method \cite{lu20172} to extract the dominant directions from the image. The dominant direction extraction is done only once before training.

Both the extracted directions and their reverse directions are considered to be the possible normal directions of the major planes in the scene, such as the ceiling, the floor, and the walls.

\textbf{Surface normal estimation.}
To estimate the surface normal, we first get the 3D coordinates $\bv{X}_p \in \mathbb{R}^{3}$ of each pixel $\bv{p}$ from the predicted depth by
\beq
\bv{X}_p = D(\bv{p}) \ba{K}^{-1} \bv{p}.
\eeq
Here, $D(\bv{p})$ denotes the depth predicted by the depth network. Next, we adopt a differentiable point-to-normal layer\cite{yang2018lego,yang2018unsupervised,kaneko2019tridepth} to estimate the surface normal from the 3D points.
Specifically, the normal ${\bv{n}}_{p}$ of a given pixel $\bv{p}$ is calculated from a set of 3D points within a small neighborhood centering on point $\bv{X}_p$.  The neighborhood is set as $7 \times 7$ in our implementation as the previous work\cite{yang2018lego}.

% lego使用的是depth-to-normal,将depth转换为三维点，三维点转换为normal，本文只用第二步，第一步使用同monodepth
\textbf{Manhattan normal detection.}
Given the surface normal prediction ${\bv{n}}$, we propose the Manhattan normal detection to classify the surface normal that belongs to the dominant planes. Our strategy is to compare the difference between the estimated normal vector ${\bv{n}}_{p}$  and each dominant direction $\bv{\eta}^k$ by using a cosine similarity  $s(\cdot ,\cdot)$ and choose the one with the best similarity, namely
\beq
\bv{n}^{align}_{p} \leftarrow{ 
	\mathop{\arg\max}\limits_{\bv{\eta}^k}
	s({\bv{n}}_{p}, \bv{\eta}^k)
}
\label{eq:aligned_norm}
\eeq
where $\bv{n}^{align}_{p}$ is the aligned normal and the cosine similarity is defined as
$
s({\bv{n}}_{p}, \bv{\eta}_k) = ({\bv{n}}_{p} \cdot \bv{\eta}_k) / 
(\| {\bv{n}}_{p}\| \cdot \|\bv{\eta}_k \|).
$
Let the maximum similarity of each pixel be $s^{max}_{p}$. We define the Manhattan mask as:
\beq
\mathcal{M}^M_{p}=\left\{
\begin{array}{rcl}
	1   &      & {s^{max}_{p} \geq \gamma}\\
	0   &      & {s^{max}_{p} < \gamma}
\end{array} \right.
\eeq where $1$ and $0$ represent Manhattan and non-Manhattan regions respectively

During the training, we use an adaptive thresholding scheme for detecting the Manhattan regions. We initially set a relatively small threshold to allow 
more pixels being classified into the Manhattan region because of inaccurate normal estimates, and gradually increase the threshold since the normal estimates become accurate in later epochs.
In our implmentation, the threshold $\gamma$ grows with the iteration number $N^{train}$ linearly: $\gamma = \alpha \cdot N^{train} + \beta$, where $\alpha$ and $\beta$ are set to $1.633 e^{-3}$ and $0.9$ respectively.

{ \bf Manhattan normal loss.  }
We apply the Manhattan normal constraint 
within the Manhattan region by using the aligned normal obtained in \Eq{eq:aligned_norm} as the supervisory signal. 
The constraint enforces the estimated normal to be as close to the aligned normal as possible, which is described by a  loss function $L_{norm}$ :
\beq
L_{norm} = \frac{1}{N_{norm}} \sum_{p} \mathcal{M}^M_{p} \mathcal{M}^P_p(1-s({\bv{n}}_{p},\bv{n}^{align}_{p}))	% \ba{M}^{plane}_{p}
\eeq where $N_{norm}$ is the number of pixels located in Manhattan regions, and $\mathcal{M}^{P}_p$ indicates whether the pixel $p$ locates in the planar regions, which we'll introduce how to detect them in the following section.

\subsection{Co-planar constraint}
% \label{sec:Co-planar constraint}

% figure2. Plane segmentation flowchart
\begin{figure}
	\includegraphics[width=0.48\textwidth]{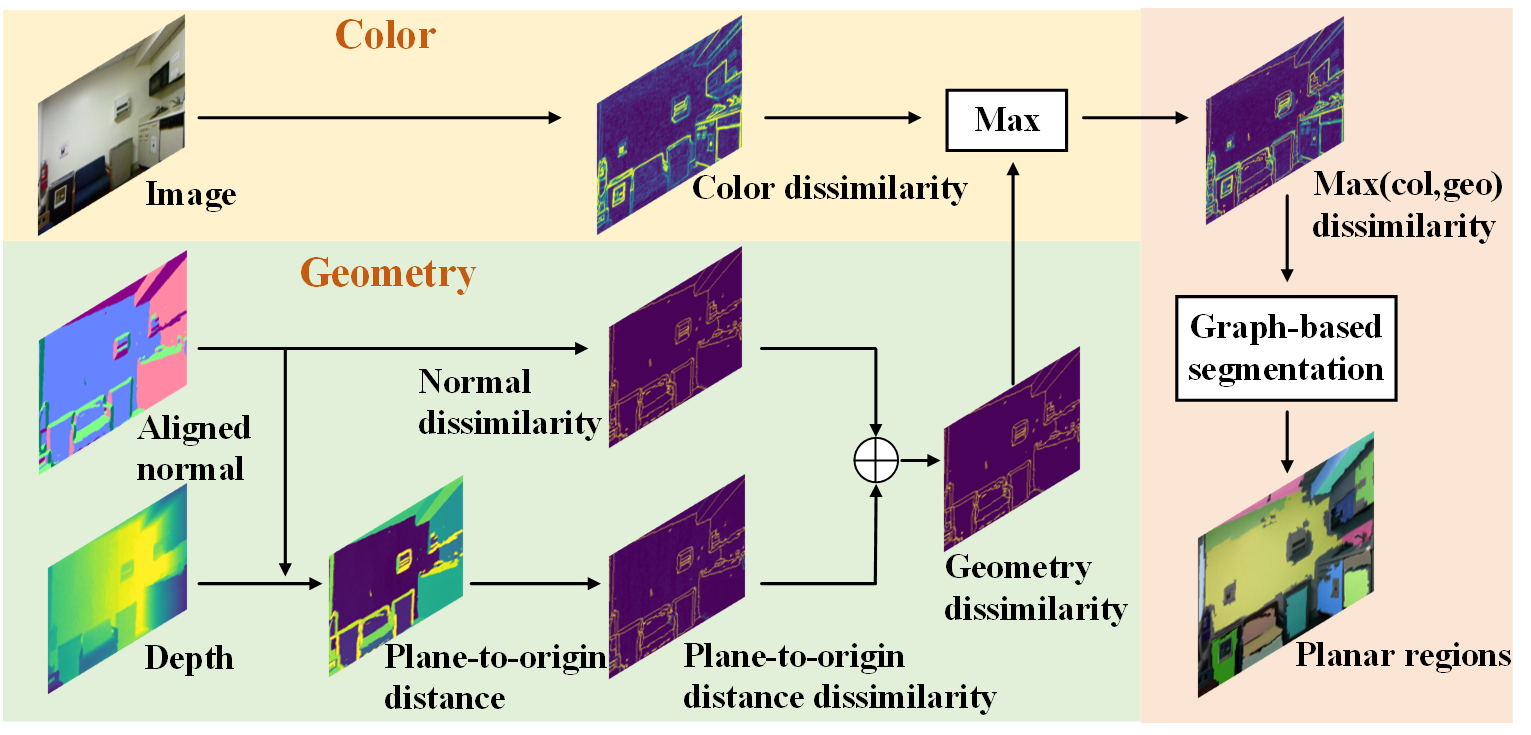}	% 0.45
	\caption{The pipeline of planar region detection.
		Both the color and geometric information are used to compute the dissimilarity for planar region segmentation.
		The color dissimilarity is calculated by comparing the RGB colors.
		The geometry dissimilarity is the sum of the normal and the plane-to-origin distance dissimilarities.
		Based on the proposed dissimilarity,  a graph-based segmentation \cite{felzenszwalb2004efficient} is applied to extract the planar regions.
	}
	\label{fig:seg_flowchart}
%%% \vspace{-0.3cm} 
\end{figure}

% fig4. examples of difficult segmentation
\begin{figure*}[ht]
	\centering
	\includegraphics[width=0.97\textwidth]{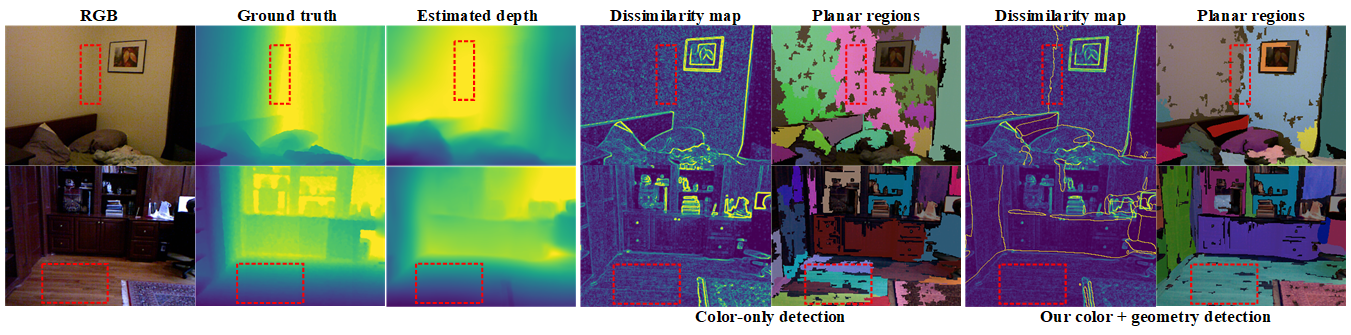} % 0.98
	\caption{The proposed planar region detection during training.
		From the left to the right columns: the input images, the groud-truth depth, the estimated depth, the dissimilarity map, and the planar regions detected by only colors \cite{yu2020p} and our method based on the color and geometric information.
		\textbf{First row}: Two walls cannot be distinguished by colors, but can be separated by our method.
		\textbf{Second row:} The floor is over-segmented by using only colors but can be correctly detected by our method.
	}
	\label{fig:method_seg_compare}
	%		\vspace{-0.4cm} 
\end{figure*}

\textbf{Planar region detection.}
To enforce the co-planar constraint, we need to detect the piece-wise planar region correctly. Previous work \cite{yu2020p} detects the planar regions by assuming the regions with homogeneous colors are planar. This simple strategy, however, usually leads to false detection or over-segmentation producing false supervisory signals. We propose a novel planar region detection method, as shown in \Fig{fig:seg_flowchart}, which integrates both the color and the online updated geometry information to extract the planar areas more reliably. 

The key idea is that we adopt a novel dissimilarity map in the following graph-based segmentation. This dissimilarity takes the color, normal, and the plane-to-origin distance into consideration.  We use the aligned normal to derive the dissimilarity instead of the estimated normal since we found the latter is too noisy.
Let the 3D coordinates of a pixel $\bv{p}$ to be $\bv{X}_p$. Suppose this 3D point lies in the plane where the normal is the aligned normal $\bv{n}^{align}_p$. The plane-to-origin distance is computed as 
\beq
d_p =  -\bv{X}_p^\vts\bv{n}^{align}_p.
\eeq
Let $q$ be the adjacent pixel of $p$. The normal dissimilarity between them is defined as the Euclidean distance between the two vectors:
\beq
\mathcal{D}_n(p,q) = \|\bv{n}^{align}_p - \bv{n}^{align}_q\|.
\eeq
Denoting the minimum and maximum dissimilarities among all the adjacent pixels by $D_n^{max}, D_n^{min}$ respectively, we define a $[\cdot]$ operator to normalize the dissimilarity via
\beq
[\mathcal{D}_n(p,q)] = (\mathcal{D}_n(p,q)- D_n^{min})/(\mathcal{D}_n^{max}-\mathcal{D}_n^{min}).
\eeq
The plane-to-origin distance dissimilarity is defined as
\beq
\mathcal{D}_d(p,q) = |d_p - d_q|.
\eeq
The geometric dissimilarity combines the normalized version of the two dissimilarities as 
\beq
\mathcal{D}_g(p,q) = [\mathcal{D}_n(p,q)] +[\mathcal{D}_d(p,q)]. 
\eeq
The color dissimilarity is computed as
\beq
\mathcal{D}_c(p,q) = \|\bv{I}_p - \bv{I}_q\|,
\eeq
where $\bv{I}_p,\bv{I}_q$ are the RGB colors.
Finally, we get the dissimilarity combining both the color and geometric information by
\beq
\mathcal{D}(p,q)  = \max([\mathcal{D}_c(p,q)],[\mathcal{D}_g(p,q)]).
\eeq
% felz method
Based on the dissimilarity, we apply the graph-based segmentation \cite{felzenszwalb2004efficient} and
filter out small areas to obtain the planar regions following \cite{yu2020p}. The advantage of using such a dissimilarity definition can be seen in \Fig{fig:method_seg_compare}. Comparing with using only the color information, our method avoids false planar regions that cannot be distinguished by colors and also over-segmentation caused by different colors.

Note that our planar region segmentation is be updated during training. As the training progresses, the gradually improved depth leads to better segmentation and vise versa.

\begin{figure*}[ht]
	\includegraphics[width=0.99\textwidth]{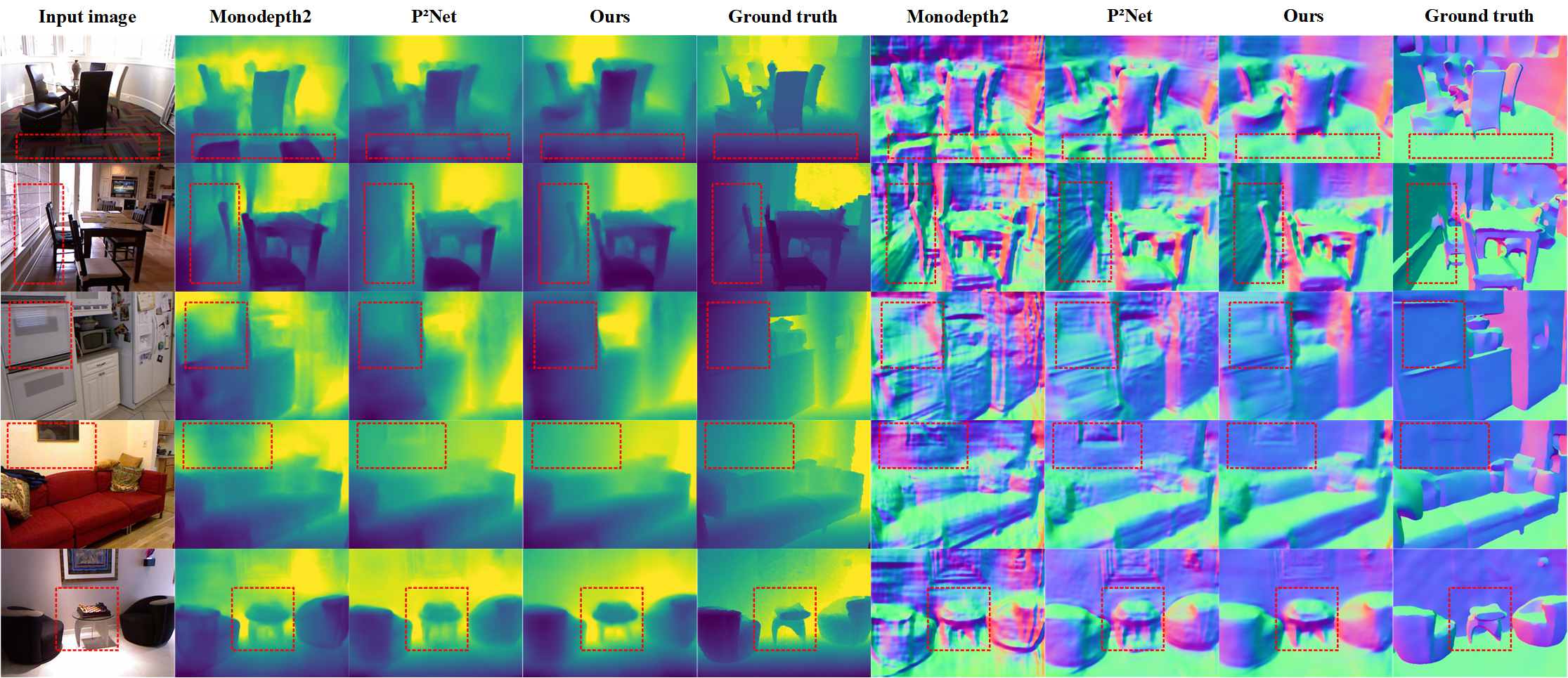}	% 0.99
	\caption{Visualization of the NYUv2 results, better viewed by zooming on screen. The depth results are on the left columns, and the surface normal results are on the right columns. The results of Monodepth2\cite{godard2019digging}, P$^{2}$Net\cite{yu2020p}, and the ground-truth depth / normal are presented for comparison. Compared with P$^2$Net\cite{yu2020p} and Monodepth2\cite{godard2019digging}, our method obtains better surface normal and depth estimation as indicated by the red rectangles. Please refer to the \Tab{tab:nyuv2 depth} and \Tab{tab:nyuv2 norm} for the quantitative results.} 
	
	\label{Fig:norm results visualization}
\end{figure*}

% table 1
\begin{table*}[ht]
	\centering
	\begin{varwidth}[t]{\textwidth}
\vspace{0pt}
		\small
		\begin{tabularx}{0.67\textwidth}{|l|c|XXX|XXX|}
			\hline
			Method & Sup. & RMS$\downarrow$ & AbsRel$\downarrow$ & Log10$\downarrow$ & $\delta_1\uparrow$ & $\delta_2\uparrow$ & $\delta_3\uparrow$ \\
			\hline
			%Li et al.(2017)\cite{li2017two} & $\surd$     & 0.635 & 0.143 & 0.063 & 78.8  & 95.8  & 99.1 \\
			%Xu et al.(2017)\cite{xu2017multi} & $\surd$ & 0.586 & 0.121 & 0.052 & 81.1  & 95.4  & 98.7 \\
			%DORN(2018)\cite{fu2018deep}  & $\surd$     & 0.509 & 0.115 & 0.051 & 82.8  & 96.5  & 99.2 \\
			Hu et al.(2019)\cite{hu2019revisiting} & $\surd$     & 0.530  & 0.115 & 0.050  & 86.6  & 97.5  & 99.3 \\
			Yin et al.(2019)\cite{yin2019enforcing}& $\surd$     & 0.416  & 0.108 & 0.048  & 87.5  & 97.6  & 99.4 \\
			AdaBins(2021)\cite{bhat2021adabins} & $\surd$        & 0.364  & 0.103 & 0.044  & 90.3  & 98.4  & 99.7 \\
			Niklaus et al.(2019)\cite{niklaus20193d} & $\surd$   & 0.300  & 0.080 & 0.030  & 94.0  & 99.0  & 100.0 \\
			\hline
			PlaneNet(2018)\cite{liu2018planenet} & $\surd$    & 0.514 & 0.142 & 0.060  & 81.2  & 95.7  & 98.9 \\
			PlaneReg(2019)\cite{yu2019single} & $\surd$    & 0.503 & 0.134 & 0.057 & 82.7  & 96.3  & 99.0 \\
			\hline
			MovingIndoor(2019)\cite{zhou2019moving} & $\times$     & 0.712 & 0.208 & 0.086 & 67.4  & 90.0    & 96.8 \\
			%Monodepth2(2019)\cite{godard2019digging}   & $\times$    & 0.600 & 0.161  & 0.068 & 77.0  & 94.8  & 98.7 \\
			Monodepth2(2019)\cite{godard2019digging}   & $\times$    & 0.600 & 0.161  & 0.068 & 77.1  & 94.8  & 98.7 \\
			$\mathrm{P}^{2}\mathrm{Net}$(2020)\cite{yu2020p} & $\times$     & 0.561 & 0.150  & 0.064 & 79.6  & 94.8  & 98.6 \\
			% $\mathrm{P}^{2}\mathrm{Net}$-finetune & $\times$     & 0.555 & 0.147  & 0.062 & 80.4  & 95.2  & 98.7 \\
			%\textbf{Ours}   & $\times$    & \textbf{0.541} & \textbf{0.142} & 
			%\textbf{0.061} & \textbf{81.2}  & \textbf{95.5}  & \textbf{98.8} \\
			\textbf{Ours}   & $\times$    & \textbf{0.540} & \textbf{0.142} & 
			\textbf{0.060} & \textbf{81.3}  & \textbf{95.4}  & \textbf{98.8} \\
			
			%\textbf{Ours $+$ pp} & $\times$     & \textbf{0.534} & \textbf{0.140} & \textbf{0.060} & \textbf{81.6} & \textbf{95.6} & \textbf{98.8} \\
			\textbf{Ours $+$ pp} & $\times$     & \textbf{0.534} & \textbf{0.140} & \textbf{0.060} & \textbf{81.7} & \textbf{95.5} & \textbf{98.8} \\
			\hline
			%\textbf{Ours$^{\star}$ }   & $\times$     & \textbf{0.536} & \textbf{0.141} & \textbf{0.060}  & \textbf{81.7}  & \textbf{95.5}  & \textbf{98.8} \\
			% \textbf{Ours$^{\star}$ }   & $\times$     & 0.536 & 0.141 & \textbf{0.060}  & \textbf{81.7}  & \textbf{95.5}  & \textbf{98.8} \\
			% \textbf{Ours$^{\star}$ $+$ pp} & $\times$     & \textbf{0.529} & \textbf{0.139} & \textbf{0.059} & \textbf{82.0} & \textbf{95.6} & \textbf{98.9} \\
			% \hline
		\end{tabularx}
	\end{varwidth}
	\quad
	\begin{varwidth}[t]{\textwidth}
\vspace{0pt}
		\parbox[b]{0.30\textwidth}{
			\small
			The first two blocks list the results of supervised methods. The second block contains the supervised methods with plane detection. The third and fourth blocks list the results of self-supervised methods. $\downarrow$ indicates the lower the better, $\uparrow$ indicates the higher the better. Our approach performs best among the self-supervised ones.\\
			
			$\surd$ - supervised learning\\
			$\times$ - self-supervised learning\\
			\textbf{pp} - with post processing as in \cite{godard2017unsupervised}\\
			% $^{\star}$ - augmentation by flipping %flipping all images for training
		}
	\end{varwidth}
	\newline
	\caption{Depth estimation results on NYUv2 dataset. }
	\label{tab:nyuv2 depth}
	%\vspace{-0.3cm} 
\end{table*}

% $\mathrm{P}^{2}\mathrm{Net}$-ft

\textbf{Generate the co-planar depth.} After detection of planar regions, we invoke the co-planar constraint to flatten the 3D points located within those plane regions. The first step is plane fitting for 3D points within the planar region. We obtain the plane parameters $\bv{\theta} = -\bv{n}/d \in\mathbb{R}^{3}$ as previous work\cite{li2020textslam,yu2020p} by solving the least squares problem
\beq
\bv{X}^\vts\bv{\theta}  = \bv{1},
\eeq
where each column of $\bv{X} \in \mathbb{R}^{3 \times N}$ represents a 3D point within the planar region. 
After that, the inverse depth $\rho_{p}$ of the pixel $\bv{p}$ by plane fitting is computed as
\beq
\rho^{plane}_{p}  =  \bv{\theta}^\vts \ba{K}^{-1} \bv{p} = 1/D^{plane}_{p},
\eeq
where $\ba{K}$ represents the camera intrinsic matrix. We then transform the inverse depth to the depth $D^{plane}_{p}$ with the maximum and minimum protection following \cite{godard2017unsupervised,godard2019digging,yu2020p}.

\textbf{Co-planar loss.} The depth $D^{plane}_p$ obtained from plane fitting is then used as an extra signal to constrain the estimated depth. The loss function is defined as 
\beq
L_{plane} = \frac{1}{N_{plane}} \sum_{p} \mathcal{M}^{P}_p \left|D_{p}-D^{plane}_{p} \right|,
\eeq where $N_{plane}$ is the number of pixels within the planar regions $\mathcal{M}^{P}$. 

\subsection{Total loss}

We use the image patches instead of individual pixels to compute the photometric loss as suggested in \cite{yu2020p}, which is defined as the combination of L1 loss and a structure similarity loss SSIM\cite{SSIM}:
\beq
\begin{aligned}
	&L_{photo} = \omega L_{SSIM} + (1-\omega) \Vert I_{t}[\mathcal{N}_{p}^{t}]-I_{s}[\mathcal{N}_{p}^{t\rightarrow s}] \Vert_{1} \\
	&L_{SSIM} = SSIM(I_{t}[\mathcal{N}_{p}^{t}],I_{s}[\mathcal{N}_{p}^{t\rightarrow s}]) 
\end{aligned}
\eeq where $\mathcal{N}_{p}$ denotes the local window surrounding $\bv{p}$. $\omega$ is the relative weight of two parts and set as $0.85$ the same as previous work\cite{godard2019digging}. We also adopt the edge-aware smoothness loss
\beq
%\vspace{-0.13cm} 
L_{smooth} = \left|\partial_x \rho_{t} \right|e^{-\left|\partial_x I_t\right|} +
\left|\partial_y \rho_{t} \right|e^{-\left|\partial_y I_t\right|},
%\vspace{-0.1cm} 
\eeq where $\rho_{t} \leftarrow \rho_{t} / \overline{\rho_{t}}$ is the mean-normalized inverse depth, 
and $\partial_x$, $\partial_y$ are the gradients along the $x$ and $y$ directions. 
The overall loss is defined as
\beq
%\vspace{-0.13cm} 
L = L_{photo} + \lambda_1 L_{smooth} + \lambda_2 L_{norm} + \lambda_3 L_{plane},
%\vspace{-0.1cm} 
\eeq
where $\lambda_1$, $\lambda_2$ and $\lambda_3$ are set to 0.001, 0.05, 0.1, respectively.

% table 2
\begin{table}
	% \small
	\scriptsize
	\centering
	\begin{tabularx}{0.48\textwidth}{|l|X|X|XXX|}
		\hline
		Method & \multicolumn{1}{l|}{Train} & \multicolumn{1}{l|}{Mean$\downarrow$} & 11.2$^{\circ}\uparrow$ & 22.5$^{\circ}\uparrow$ & 30$^{\circ}\uparrow$ \\
		\hline\hline
		\multicolumn{6}{|c|}{Surface normal estimation networks} \\
		\hline
		3DP(2013)\cite{fouhey2013data}   & $\surd$& 33.0 & 18.8 & 40.7 & 52.4 \\ 
		Fouhey et al.(2014)\cite{fouhey2014unfolding} & $\surd$ & 35.2 & 40.5 & 54.1 & 58.9 \\
		%Wang et al.(2014)\cite{wang2015designing} & $\surd$ & 28.8  & 35.2 & 57.1  & 65.5 \\
		Wang et al.(2015)\cite{wang2015designing} & $\surd$ & 28.8  & 35.2 & 57.1  & 65.5 \\
		Eigen et al.(2015)\cite{eigen2015predicting} & $\surd$ & 23.7 & 39.2 & 62.0  & 71.1 \\
		\hline\hline
		\multicolumn{6}{|c|}{Surface normal computed from the depth} \\
		\hline
		GeoNet(2018)\cite{qi2018geonet} & $\surd$ & 36.8 & 15.0 & 34.5 & 46.7 \\
		DORN(2018)\cite{fu2018deep}  & $\surd$ & 36.6 & 15.7 & 36.5 & 49.4 \\
		\hline
		MovingIndoor(2019)\cite{zhou2019moving} & $\times$ & 43.5 & 10.2 & 26.8 & 37.9 \\
		Monodepth2(2019)\cite{godard2019digging} &$\times$ & 45.1 & 10.4 & 27.3 & 37.6 \\
		$\mathrm{P}^{2}\mathrm{Net}$(2020)\cite{yu2020p} & $\times$ & 36.6 & 15.0 & 36.7 & 49.0 \\
		% $\mathrm{P}^{2}\mathrm{Net}$-finetune & $\times$ & 37.1 & 15.6 & 36.7 & 48.8 \\
		%\textbf{Ours} & $\times$ & \textbf{34.3}  & \textbf{21.2}  & \textbf{44.0}  & \textbf{55.0} \\
		%\textbf{Ours} & $\times$ & \textbf{34.5}  & \textbf{20.9}  & \textbf{43.5}  & \textbf{54.6} \\
		\textbf{Ours} & $\times$ & \textbf{34.5}  & \textbf{21.9}  & \textbf{44.4}  & \textbf{55.2} \\
		%\textbf{Ours $+$ pp} & $\times$ & \textbf{34.1}  & \textbf{21.8}  & \textbf{44.4}  & \textbf{55.3} \\
		%\textbf{Ours $+$ pp} & $\times$ & \textbf{34.3}  & \textbf{21.5}  & \textbf{44.0}  & \textbf{54.9} \\
		\textbf{Ours $+$ pp} & $\times$ & \textbf{34.2}  & \textbf{22.6}  & \textbf{44.7}  & \textbf{55.4} \\
		\hline
		%\textcolor{red}{\textbf{Ours$^\star$}} & $\times$ & 34.7  & 22.1  & 44.3  & 55.1 \\
		%\textcolor{red}{\textbf{Ours$^\star$ $+$ pp}} & $\times$ & 34.5  & \textbf{22.9}  & \textbf{44.7}  & 55.3 \\
		%\hline
	\end{tabularx}
	\newline
	\caption{Surface normal estimation results on NYUv2. We report the results of surface normal estimation networks in the first block. The normal results computed from the depth networks are in the second and the third block, where '$\surd$' denotes supervised methods, and '$\times$' denotes self-supervised ones. 
		The normal computation is the same for all methods.  Our method outperforms existing monocular depth estimation methods in surface normal estimation.}
	\label{tab:nyuv2 norm}
\end{table}

%%% \vspace{-0.15cm} 
\section{Experimental results}
%%% \vspace{-0.1cm} 

We train our model on the NYUv2 dataset \cite{silberman2012indoor} using the data split the same as the previous work \cite{zhou2019moving}\cite{yu2020p}, and evaluate our methods on NYUv2\cite{silberman2012indoor}, ScanNet\cite{dai2017scannet}, and InteriorNet\cite{li2018interiornet} datasets.  We detect the vanishing points on the training images and skip 18 image sequences that fail to detect valid vanishing points. This results in 21465 monocular training sequences and 654 images for validation. Each monocular training sequence consists of five frames. Our network model adopts the same architecture as \cite{yu2020p}.

We compare our method with the state-of-the-art methods of monocular depth estimation. Apart from depth estimation, we also evaluate the performance of surface normal estimation, and present ablation studies about the effectiveness of the proposed supervisory signals, and using different network architectures. More results can be found in the supplementary material.

%\vspace{-0.1cm}
\subsection{Implementation details}
%\vspace{-0.1cm}
The network is trained for a total of 50 epochs with a batch size of 32 based on the pre-trained model \cite{yu2020p}. We use Adam optimizer and a multi-step learning rate reduction strategy. We set the initial learning rate as $10^{-4}$, then decay it by 0.1 at the 26th epoch and 36th epoch. We perform random flipping and color augmentation during training. All images are firstly undistorted and cropped by 16 pixels from the border, and then scaled to $288\times384$ for training. The camera intrinsic parameters come from the official specification \cite{silberman2012indoor}, and are adjusted to be consistent with the image cropping and scaling. We follow the same criteria used in \cite{godard2019digging,yu2020p}  for evaluation. Namely, we cap the depth to $10m$  and use the median scaling strategy to avoid the scale ambiguity of monocular depth estimation. The evaluation metrics include root mean squared error (RMS), absolute relative error (AbsRel), mean log10 error (Log10), and the accuracy under threshold $(\delta_i < 1.25^i, i = 1, 2, 3)$.

% \vspace{-0.1cm}
\subsection{Results on NYUv2 Dataset}
%\vspace{-0.1cm}
%We use \cite{yu2020p} as the baseline and follow their data split.
\textbf{Depth estimation.} The quantitative results of depth estimation are listed in \Tab{tab:nyuv2 depth}. The results show that our method outperforms MovingIndoor\cite{zhou2019moving} and P$^{2}$Net\cite{yu2020p}, the state-of-the-art self-supervised methods on indoor monocular depth estimation, by a large margin. The results also show that our method surpasses some supervised approaches. The depth estimation results are visualized in \Fig{Fig:norm results visualization}. We can see that our method obtains more accurate indoor structures and smoother planes than existing methods.

\textbf{Surface normal estimation.} We also evaluate the surface normal estimation as shown in \Tab{tab:nyuv2 norm}. Our method  outperforms existing methods, and also some supervised methods\cite{fouhey2013data,qi2018geonet,fu2018deep}. Results are also shown in \Fig{Fig:norm results visualization}.	% Our method  outperforms the baseline method, 

%\vspace{-0.1cm}
\subsection{Results on ScanNet and InteriorNet}
%\vspace{-0.1cm}

% fig6
\begin{figure}[ht]
	\centering
	\includegraphics[width=0.48\textwidth]{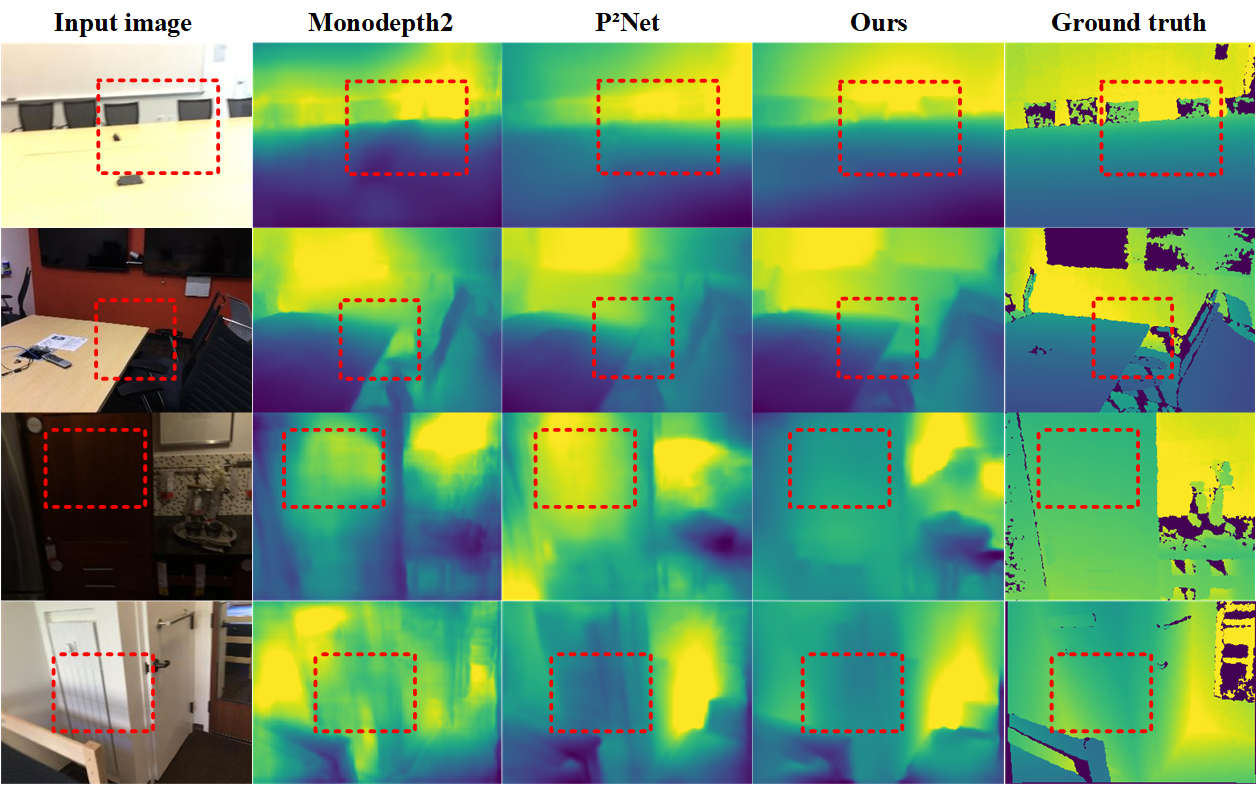} % 0.48
	\caption{\textbf{ScanNet results} with the trained model on NYUv2. The holes in the ground truth are excluded from evaluation.} % , illustrating better structures.
	\label{Fig:scannet depth}
\end{figure}
% Tab3
\begin{table}[h]
	\scriptsize                                                                                    
	%\small
	\centering
	\begin{tabularx}{0.48\textwidth}{|l|XXX|XXX|}
		\hline
		Method & RMS$\downarrow$ & AbsRel$\downarrow$ & Log10$\downarrow$ & $\delta_{1}\uparrow$ & $\delta_{2}\uparrow$ & $\delta_{3}\uparrow$ \\
		\hline
		Monov2\cite{godard2019digging} & 0.451 & 0.191   & 0.080 & 69.3  & 92.6  & 98.3 \\
		
		P$^2$Net \cite{yu2020p} & 0.420  & 0.175 & 0.074 & 74.0    & 93.2  & 98.2 \\
		
		P$^2$Net-finetune & 0.412  & 0.172 & 0.073 & 74.3    & 93.5  & 98.4 \\
		
		%\textbf{Our}   & \textbf{0.403} & \textbf{0.167} & \textbf{0.071} & \textbf{75.3}  & \textbf{93.7}  & \textbf{98.5} \\
		%\textbf{Our}   & \textbf{0.402} & \textbf{0.167} & \textbf{0.071} & \textbf{75.3}  & \textbf{93.8}  & \textbf{98.5} \\
		\textbf{Our}   & \textbf{0.400} & \textbf{0.165} & \textbf{0.070} & \textbf{75.4}  & \textbf{93.9}  & \textbf{98.5} \\
		\hline
	\end{tabularx}%
	\newline
	\caption{\textbf{ScanNet results} with the trained model on NYUv2.}
	\label{tab:scannet}
	%	\vspace{-0.4cm} 
\end{table}%

\begin{figure}[ht]                        
	\centering                                          
	\includegraphics[width=0.48\textwidth]{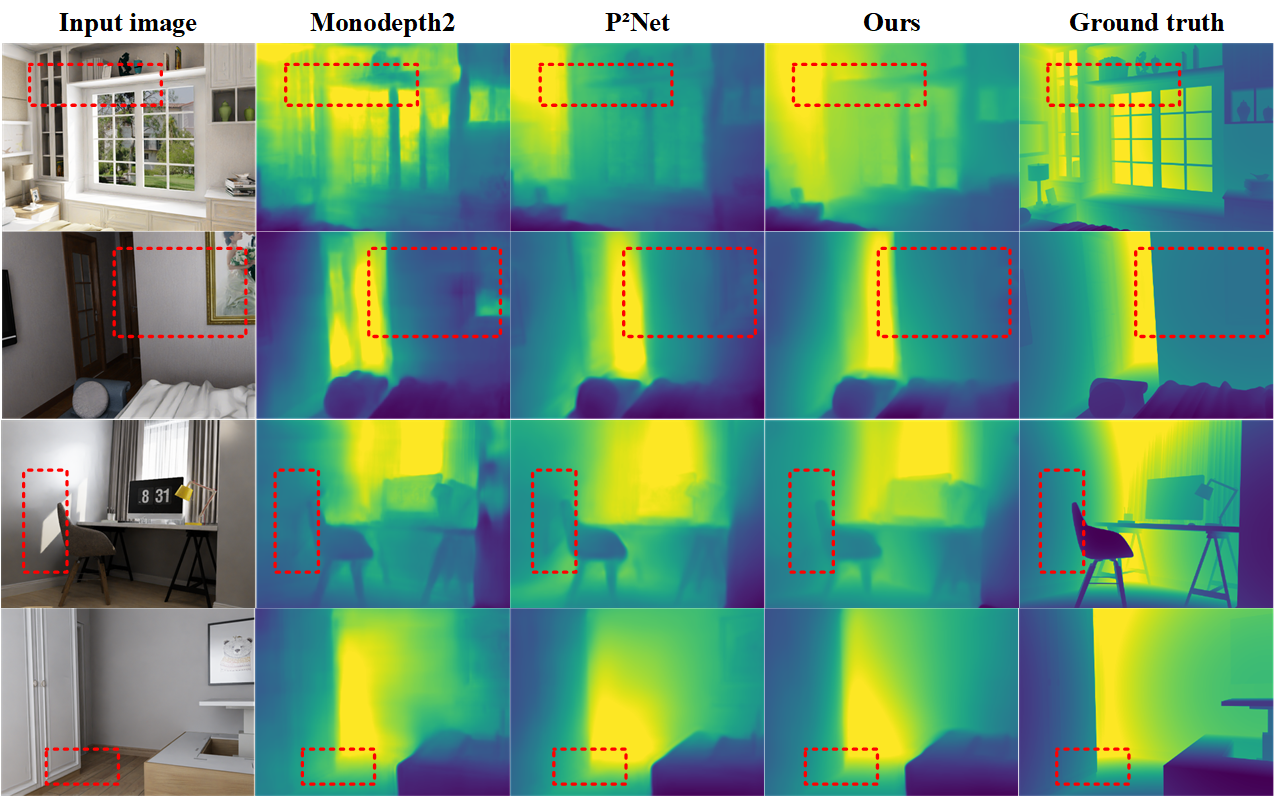} % 0.48
	\caption{\textbf{InteriorNet results} with the trained model on NYU V2.}	% , illustrating better structures
	\label{Fig:interior depth}
	%%% \vspace{-0.4cm}
	% \setlength{\abovecaptionskip}{0.cm}
	% \setlength{\belowcaptionskip}{-0.4cm}
\end{figure}

We use the model trained only on NYUv2 to evaluate our methods generalized to other indoor datasets.
\textbf{ScanNet}\cite{dai2017scannet} is captured with a depth camera attached to a iPad, containing around 2.5M RGBD video captured in 1513 scenes. We use the test split proposed by \cite{yu2020p} which includes 533 images. The evaluation results are shown in \Tab{tab:scannet} and \Fig{Fig:scannet depth}.
\textbf{InteriorNet}\cite{li2018interiornet} is a synthetic dataset of indoor video sequences containing millions of well-designed interior design layouts, furniture and object models. Because there is no current official train/test split on InteriorNet for depth estimation, here we selected 540 images randomly from the HD7 data of the full dataset as test images. The evaluation results are shown in \Tab{tab:interiornet} and \Fig{Fig:interior depth}.

Although ScanNet and InteriorNet have not been used for training, the results show that our method still generalizes well and outperforms existing methods.

% fig7

% Tab4
\begin{table}[h]
	\scriptsize
	\centering
	\begin{tabularx}{0.48\textwidth}{|l|XXX|XXX|}
		\hline
		Method & RMS$\downarrow$ & AbsRel$\downarrow$ & Log10$\downarrow$ & $\delta_{1}\uparrow$ & $\delta_{2}\uparrow$ & $\delta_{3}\uparrow$ \\
		\hline
		Monov2\cite{godard2019digging} & 0.817 & 0.368  & 0.124 & 58.6  & 81.5  & 89.8 \\
		
		P$^2$Net \cite{yu2020p} & 0.737  & 0.346 & 0.115 & 64.2    & 83.3  & 90.2 \\
		
		P$^2$Net-finetune  & 0.736  & 0.340 & 0.114 & 64.4    & 83.3  & 90.3 \\
		
		%\textbf{Our}   & \textbf{0.716} & \textbf{0.331} & \textbf{0.112} & \textbf{65.6}  & \textbf{83.9}  & \textbf{90.5} \\
		%\textbf{Our}   & \textbf{0.718} & \textbf{0.332} & \textbf{0.112} & \textbf{65.4}  & \textbf{83.9}  & \textbf{90.5} \\
		\textbf{Our}   & \textbf{0.715} & \textbf{0.330} & \textbf{0.111} & \textbf{66.0}  & \textbf{84.0}  & \textbf{90.5} \\
		\hline
	\end{tabularx}
	\newline
	\caption{\textbf{InteriorNet results} with the trained model on NYUv2.}
	\label{tab:interiornet}
	%%% \vspace{-0.4cm} 
\end{table}%

% tab5
\begin{table}[ht]
	\scriptsize
	\centering
	\begin{tabularx}{0.48\textwidth}{|l|XXX|XXX|}
		\hline                                                    
		Methods &  RMS$\downarrow$ & AbsRel$\downarrow$ & Log10$\downarrow$ & $\delta_1\uparrow$ & $\delta_2\uparrow$ & $\delta_3\uparrow$ \\
		\hline
		P$^2$Net\cite{yu2020p} & 0.561 & 0.150 & 0.064 & 79.6  & 94.8  & 98.6 \\
		\hline
		\hline
		P$^2$Net-finetune & 0.555 & 0.147 & 0.062 & 80.4  & 95.2  & 98.7 \\
		
		%Coplanar-only & 0.544 & 0.143  & 0.061 & 81.0  & 95.4  & 98.8 \\
		%Coplanar-only & 0.546 & 0.143  & 0.061 & 80.9  & 95.4  & 98.8 \\
		Coplanar-only & 0.548 & 0.144  & 0.061 & 80.8  & 95.3  & 98.8 \\
		
		%Normal-only & 0.551 & 0.145 & 0.061 & 80.9 & 95.4  & 98.8 \\
		Normal-only & 0.543 & 0.143 & 0.061 & 81.0 & \textbf{95.5}  & \textbf{98.9} \\
		
		%\textbf{Our(full)} & \textbf{0.540} & \textbf{0.142} & 
		%\textbf{0.061} & \textbf{81.3}  & \textbf{95.4}  & \textbf{98.8} \\
		%\textbf{Our(full)} & \textbf{0.541} & \textbf{0.142} & 
		%\textbf{0.061} & \textbf{81.2}  & \textbf{95.5}  & \textbf{98.8} \\
		\textbf{Our(full)} & \textbf{0.540} & \textbf{0.142} & 
		\textbf{0.060} & \textbf{81.3}  & 95.4  & 98.8 \\
		\hline
	\end{tabularx}%
	\newline
	\caption{Ablation study about using different supervisory signals. 
		% trained without our constraints based on the pre-trained model \cite{yu2020p}.
		% we trained P2Net model following [47] strategy with the same epoch number as ours for ablation study (the ’Baseline’ in Tab5). 
		We evaluate the performances using only the Manhattan normal constraint (Normal-only), using only the co-planar constraint (Coplanar-only), and the proposed method (Our(full)). We also present the result of fine-tuned P$^2$Net model (P$^2$Net-finetune). Note all the models were trained with the same number of epochs for fair comparison. }
	\label{tab:ablation}%
%%% \vspace{-0.3cm} 
\end{table}%

% table 6
\begin{table}
	%	\vspace{-0.8cm} 
	\scriptsize
	\centering
	\begin{tabularx}{0.48\textwidth}{|l|XXX|XXX|}
		\hline
		Train & RMS$\downarrow$ & AbsRel$\downarrow$ & Log10$\downarrow$  & $\delta_1\uparrow$ & $\delta_2\uparrow$ & $\delta_3\uparrow$ \\
		\hline
		\hline
		\multicolumn{7}{|c|}{Using the Monodepth2 \cite{godard2019digging} architecture} \\
		\hline
		Original & 0.600 & 0.161 & 0.068 & 77.1 & 94.8  & 98.7 \\
		Original-finetune & 0.598 & 0.159 & 0.067 & 77.5 & 94.9  & 98.7 \\
		\textbf{Ours}  & \textbf{0.564}  & \textbf{0.151} & \textbf{0.065} & \textbf{79.1}  & \textbf{95.0}  & \textbf{98.8} \\
		\hline
		\hline
		\multicolumn{7}{|c|}{Using the P$^{2}$Net \cite{yu2020p} architecture} \\
		\hline
		Original & 0.561 & 0.150 & 0.064 & 79.6  & 94.8  & 98.6 \\
		Original-finetune & 0.555 & 0.147  & 0.062 & 80.4  & 95.2  & 98.7 \\
		%\textbf{Ours}  & \textbf{0.540} & \textbf{0.142} & \textbf{0.061} & \textbf{81.3}  & \textbf{95.4}  & \textbf{98.8} \\
		%\textbf{Ours}  & \textbf{0.541} & \textbf{0.142} & \textbf{0.061} & \textbf{81.2}  & \textbf{95.5}  & \textbf{98.8} \\
		\textbf{Ours}  & \textbf{0.540} & \textbf{0.142} & 
		\textbf{0.060} & \textbf{81.3}  & \textbf{95.4}  & \textbf{98.8} \\
		\hline
	\end{tabularx}
	\newline
	\caption{Ablation study about using different network architectures. Our extra training losses improves both models, indicating our method is universal to different architectures. }
	\label{tab:diff baseline}  
\end{table} 

%%% \vspace{-0.2cm} 
\subsection{Ablation study}
%%% \vspace{-0.1cm} 
To better understand the effectiveness of each part of our method, we perform an ablation study by changing various components of our model on NYU V2 dataset. We initialize the network with the pre-trained model \cite{yu2020p} and train it with the proposed supervisory signals. The results are shown in \Tab{tab:ablation}. Either the Manhattan normal loss or the co-planar loss leads to depth estimations better than that of the original and the original-finetune methods. 
% Either the Manhattan normal loss or the co-planar loss leads to depth estimations better than that of the baseline method. 
Incorporating them together leads to the maximum gain in performance.

We also test our method using different network architectures. As shown in \Tab{tab:diff baseline}, using the proposed supervisory signals, both models are improved, indicating our method is universal to different network architectures. But the results based on Monodepth2 are worse than those based on P$^2$Net. This is largely due to the patch-based photometric loss that is better for texture-less regions as suggested in \cite{yu2020p}.

%%% \vspace{-0.1cm}
\subsection{Planar-region detection in training}
\begin{figure}                              
	\centering                    
	\includegraphics[width=0.42\textwidth]{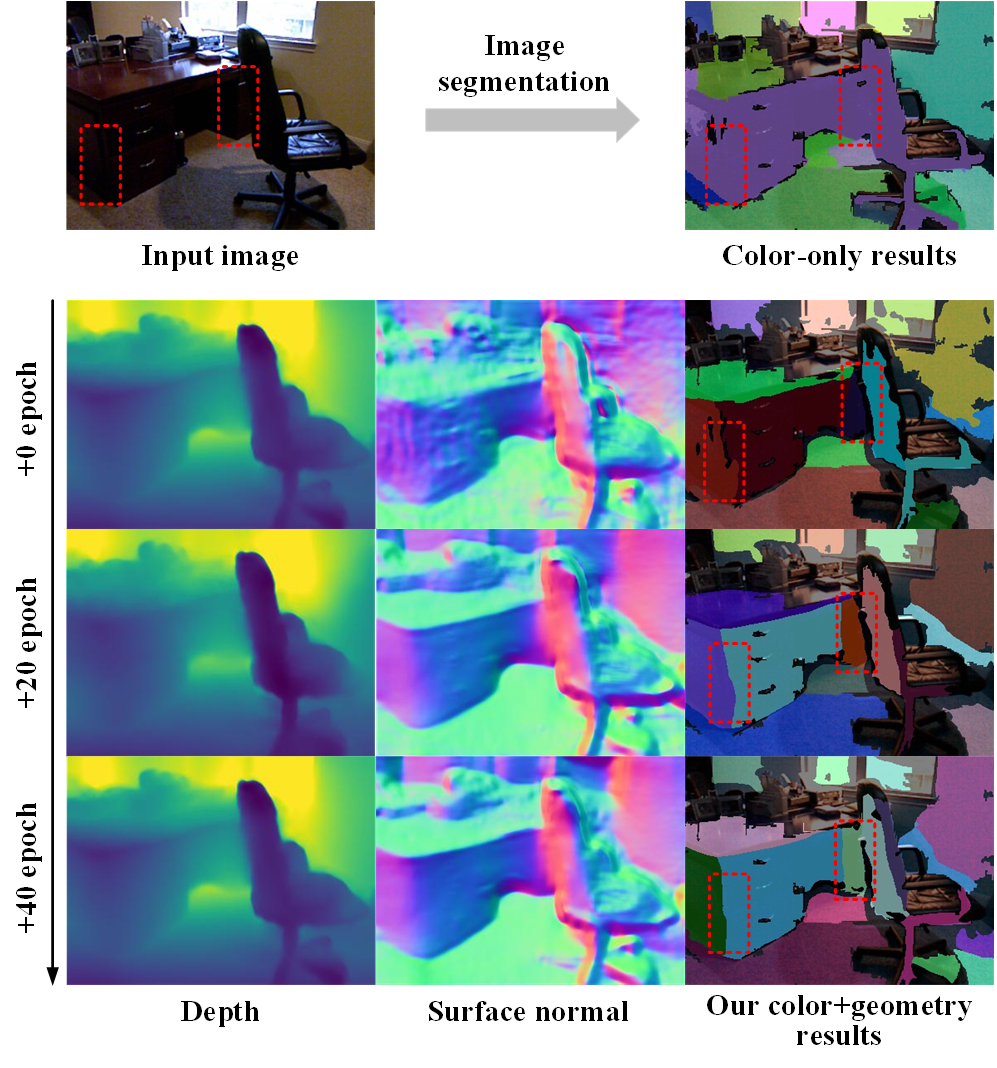}	% =0.38
	\caption{\textbf{First row}: The planar regions detected by the color-only method \cite{yu2020p}. 
		\textbf{Bottom rows}: The estimated depth, surface normal and segmentation results at different epochs on NYUv2. Our segmentation results gradually improve as the training progresses.}
	\label{Fig:seg trainning}
	%%% \vspace{-0.3cm} 
\end{figure}

%%% \vspace{-0.1cm}
We show the intermediate planar region detection results during training in \Fig{Fig:seg trainning}. 
The results show that the planar region segmentation gradually improves with the updated depth and normal estimates. 
By contrast, the color-only method produces false planar regions as indicated by the red rectangles.

%%% \vspace{-0.2cm}  
\section{Limitation}
%%% \vspace{-0.1cm} 
We discuss the limitations of our method. The first limitation is that extracting dominant directions highly relies on the Manhattan world assumption. It may not work well in indoor scenes with irregular layouts containing slant planes. Possible solutions include using a relaxed version of Manhattan world assumption as in \cite{schindler2004atlanta}\cite{zou2019structvio}, or directly using the estimated direction from each detected vanishing point to derive the normal constraint. In other words, those dominant directions are not restricted to be mutually perpendicular.
The second limitation is that the low quality of initial depth should be avoided. As our planar region detection relies on depth information, the low depth quality will deteriorate the segmentation results and generate false supervisory signals, which in turn prevent the network from converging to a good model. Our solution is to use a pre-trained depth model or train the model only with photometric and smoothness losses in early epochs. It leaves open to design a better planar region detector given low-quality initial depth estimates.

%% >[END HERE]<
%%% \vspace{-0.2cm} 
\section{Conclusion}
%%% \vspace{-0.1cm} 
In this paper, we propose to leverage the structural regularities of indoor environments for monocular depth estimation. Two extra losses, Manhattan normal loss and co-planar loss, are used to supervise the depth learning. Those supervisory signals are generated on the fly during training by Manhattan normal detection and planar region detection. Our method achieves the state-of-the-art result on indoor benchmark datasets.

{\small
	\bibliographystyle{ieee_fullname}
	\bibliography{root}
}

%%%%%%%%% PAPER END

%%%%%%%%% SUPPLYMENT BEGIN

% for SUPPLYMENTARY FORMAT
% \pagebreak[4]
\newpage
% \pagebreak
% \vfill
%\begin{widetext}

% \clearpage

\setcounter{section}{0}
\setcounter{equation}{0}
\setcounter{figure}{0}
\setcounter{table}{0}
\setcounter{page}{1}
\makeatletter
\renewcommand{\theequation}{S\arabic{equation}}
\renewcommand{\thefigure}{S\arabic{figure}}

\begin{center}
	\textbf{\large Supplementary Material}
\end{center}
%\end{widetext}

%\maketitle
%% Remove page # from the first page of camera-ready.
%\ificcvfinal\thispagestyle{empty}\fi

Here, we present extra experimental results, including the additional visual results of our method, the outdoor tests, and the plane quality tests.

% figure 0. kitti 
\begin{figure*}[h]
	\centering
	\includegraphics[width=\textwidth]{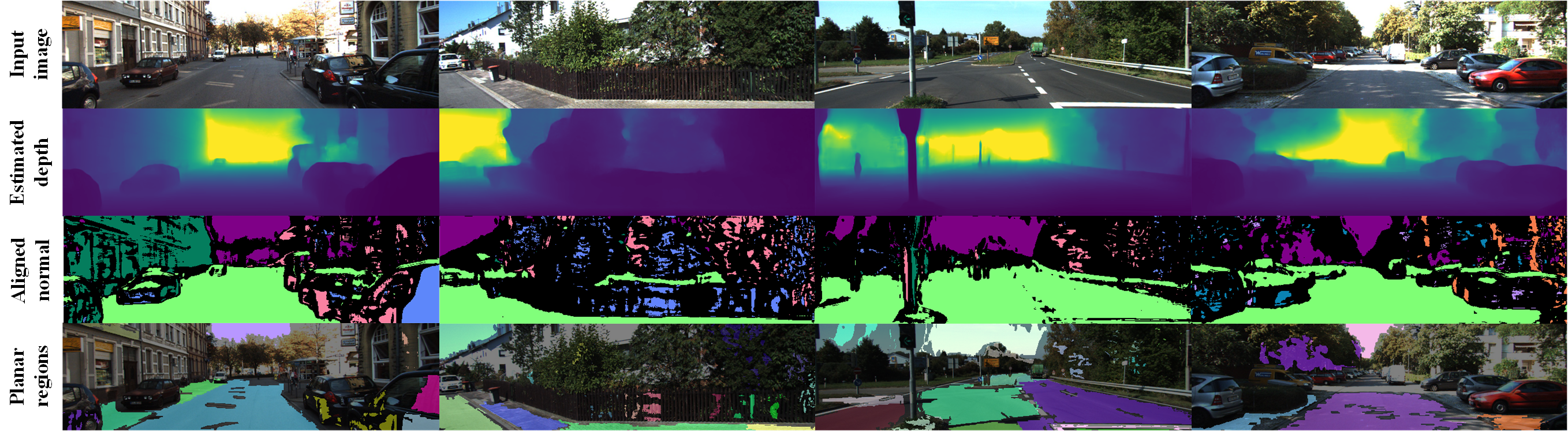}
	\caption{Visualization of the KITTI results. From top to bottom rows: the input image, the estimated depth, the aligned surface normal, and the planar regions detected by our method based on the color and geometric information.} 
	\label{Fig:kitti results visualization}
\end{figure*}

% figure 1. pointcloud
\begin{figure*}[h]
	\centering
	\includegraphics[width=0.83\textwidth]{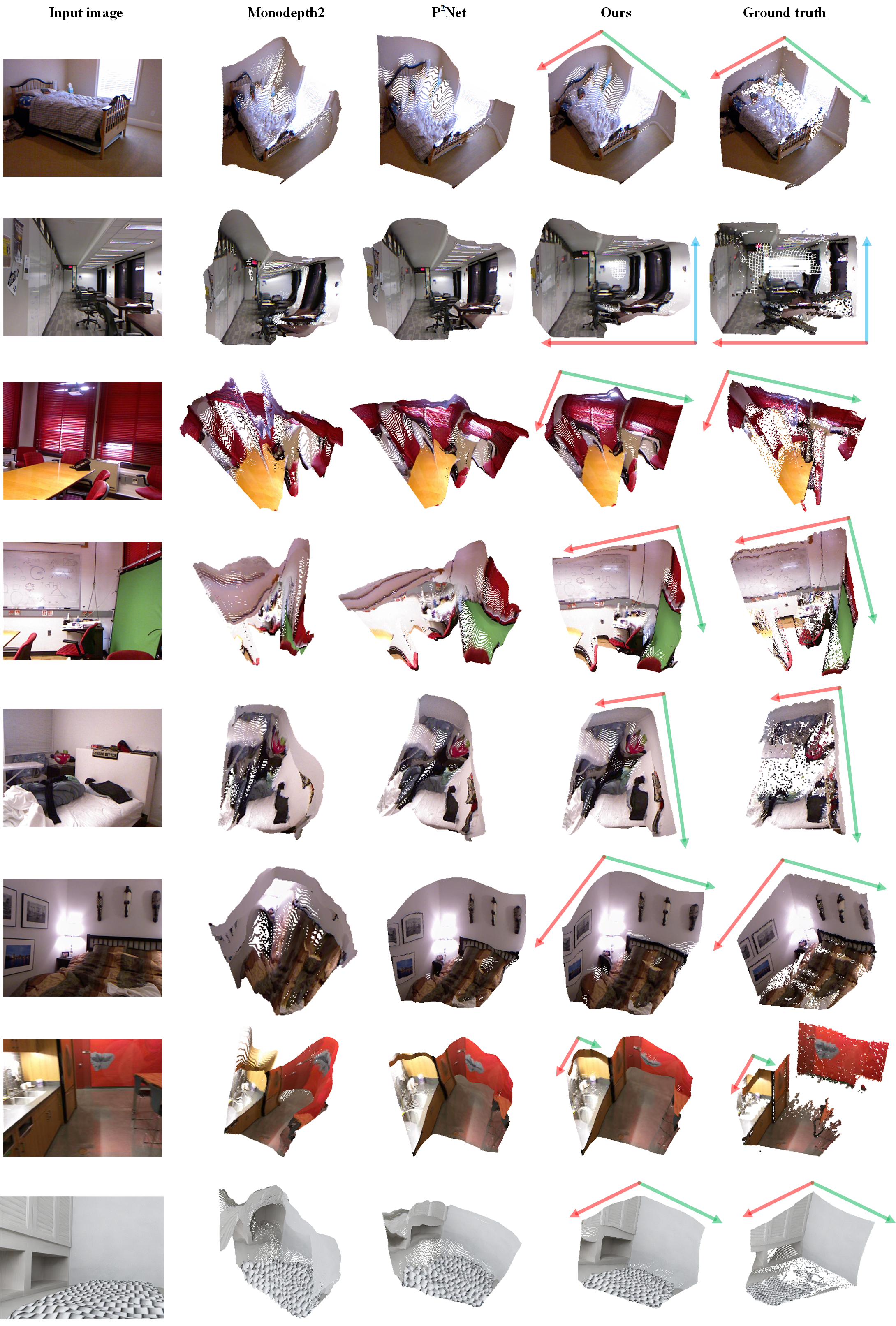}	% 0.85
	\caption{Point cloud visualization on NYUv2, ScanNet and InteriorNet results. We present the point cloud results of Monodepth2\cite{godard2019digging}, $\mathrm{P}^{2}\mathrm{Net}$\cite{yu2020p}, our method, and the ground-truth.	We draw the dominant directions in the scene for better comparison. The results show that our method produces more accurate 3D structures.
	} 
	\label{Fig:pointcloud results visualization}
\end{figure*}

\section{Extra qualitative results}
\begin{figure*}[t!]
	\includegraphics[width=0.99\textwidth]{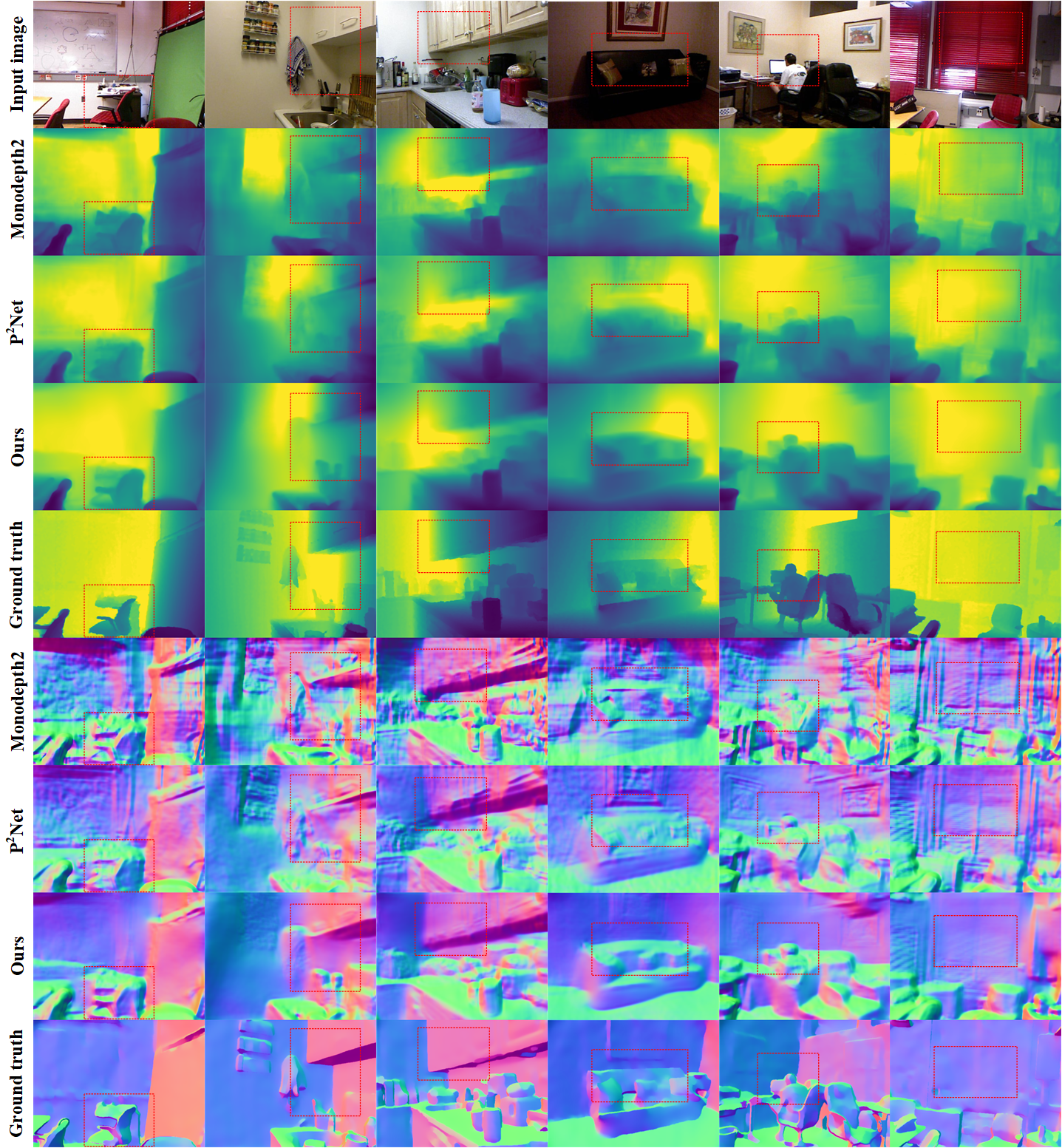}
	\caption{Qualitative visualization results on NYUv2.
 The top rows show the depth results and the bottom rows show the surface normal results.
		The results of Monodepth2\cite{godard2019digging}, $\mathrm{P}^{2}\mathrm{Net}$\cite{yu2020p}, our method, and the ground-truth depth / normal are presented for comparison.
		Compared with  $\mathrm{P}^{2}\mathrm{Net}$\cite{yu2020p} and Monodepth2\cite{godard2019digging}, our method obtains better surface normal estimation and depth prediction as indicated by the red rectangles.
		% We also 
	} 
	\label{Fig:depth results visualization nyu}
\end{figure*}

\begin{figure*}[t!]
	\includegraphics[width=0.99\textwidth]{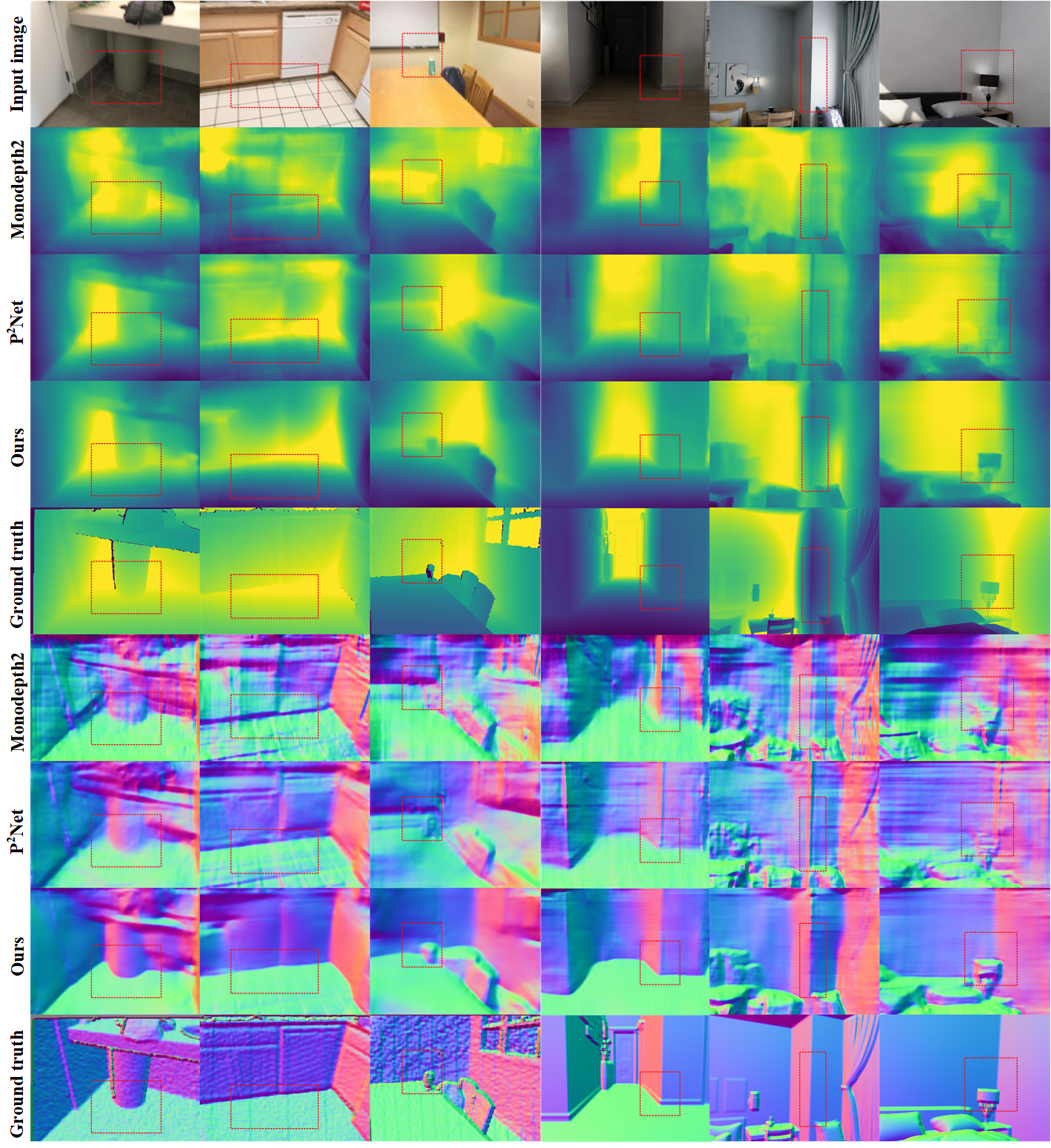}
	\caption{Qualitative visualization results on ScanNet and InteriorNet datasets.
 The top rows show the depth results and the bottom rows show the surface normal results.
		The results of Monodepth2\cite{godard2019digging}, $\mathrm{P}^{2}\mathrm{Net}$\cite{yu2020p}, ours and the ground-truth depth / normal are presented for comparison.
		Compared with  $\mathrm{P}^{2}\mathrm{Net}$\cite{yu2020p} and Monodepth2\cite{godard2019digging}, our method obtains better surface normal estimation and depth prediction as indicated by the red rectangles. Our models were trained on the NYUv2 dataset.
	} 
	\label{Fig:depth results visualization ScanInter}
\end{figure*}

We include additional qualitative results on NYUv2, ScanNet, and InteriorNet datasets.
\Fig{Fig:pointcloud results visualization} shows the 3D structure recovered from the estimated depth.
\Fig{Fig:depth results visualization nyu} and \Fig{Fig:depth results visualization ScanInter} illustrate the results of the depth and surface normal estimation.
Those results show that our method achieves more accurate depth estimation and produces more accurate 3D structures, compared with the existing methods.

% figure 2. depth norm

\section{Outdoor tests}

We present the results of our method on the KITTI dataset, which is captured in outdoor scenes.
% data
We use the split composed of 44234 images as the training dataset, the same as Monodepth2\cite{godard2019digging}.
We firstly detected the vanishing points on the training images and skipped 335 images which fail to detect valid vanishing points.
Consequently, 39500 image sequences were used for training and 4397 image sequences for validation.
Other dataset preprocessing settings are consistent with \cite{godard2019digging}. 
The total epoch number of training is 17 with a batch size of 16. The initial learning rate is $10^{-5}$ and drops to $10^{-6}$ after 15 epochs. 
Results are shown in Tab. \ref{tab:diff baseline KITTI}. 

% table 2 KITTI ablation
\begin{table}[h]
	% \small
	\scriptsize
	\centering
	\begin{tabularx}{0.48\textwidth}{|l|XXX|XXX|}
		\hline
		Train & RMS$\downarrow$ & AbsRel$\downarrow$ & SqRel$\downarrow$ & $\delta_1\uparrow$ & $\delta_2\uparrow$ & $\delta_3\uparrow$ \\
		\hline
		\hline
		\multicolumn{7}{|c|}{Using the Monodepth2 architecture} \\
		\hline
		Original & 4.863 & \textbf{0.115} & 0.903 & \textbf{87.7} & \textbf{95.9}  & \textbf{98.1} \\
		Original-finetune & 4.882 & 0.117 & \textbf{0.894} & 87.2 & 95.8  & 98.0 \\
		\textbf{Ours}  & \textbf{4.850}  & 0.120 & 0.906 & 87.0  & 95.8  & \textbf{98.1} \\
		\hline
		\hline
		\multicolumn{7}{|c|}{Using the P$^{2}$Net architecture} \\
		\hline
		Original & 5.008 & 0.121 & 0.964 &  86.6 &  95.4 & \textbf{97.9} \\
		Original-finetune & 5.041 & 0.121 & 0.996 & \textbf{86.7}  & 95.4  & 97.8  \\
		\textbf{Ours} & \textbf{4.969} & \textbf{0.120} & \textbf{0.941} & \textbf{86.7}  & \textbf{95.5}  & \textbf{97.9} \\
		\hline
	\end{tabularx}
\\
	\caption{Outdoor tests using different network architectures on KITTI dataset. }
	\label{tab:diff baseline KITTI}  
\end{table}

% results 1
From the results in \Tab{tab:diff baseline KITTI}, we can see that using
the Monodepth2 \cite{godard2019digging} architecture achieves better performance than using P$^2$Net.
This is largely due to that the outdoor environments are full of textures. The well-designed Monodepth2 works well in such kinds of scenes, while the strategies adopted in P$^2$Net are more suitable for indoor scenes. This has been discussed in \cite{yu2020p}. Though our method does not improve the performance too much by using Monodepth2 architecture, we can see the effectiveness of our extra structural losses by using the P$^2$Net architecture.

The depth, the surface normal, and the plane detection results of our method on KITTI dataset are shown in Fig. \ref{Fig:kitti results visualization}. Note that the detected planar regions are mostly located on the road, where the textures are rich enough to supervise a good depth. This may be the major reason why our extra losses did not help too much within the Monodepth2 training pipeline. The other reason may be that the extracted dominant directions may not be strictly mutually perpendicular in outdoor scenes, leading to large surface normal errors.

\section{Plane quality tests}
We evaluate the plane quality on the IBims-1\cite{koch2018evaluation} as \cite{yin2021learning}. All models are trained on the NYUv2 dataset with the same number of epochs for fair comparison (pretrain epochs are also included in our method). From the results in \Tab{tab:IBIMS}, as expected, our method produces the best plane quality (the second column), even though P2Net also adopts co-planar losses. The improvements are largely due to the global constraint from Manhattan normal loss. However, all methods produce low structure quality especially the depth edge comparing with supervised methods, indicating great efforts are still required to improve self-supervised depth learning in indoor scenes.
\begin{table}[h]
	\scriptsize                                                                                    
	%\small
	\centering
	\begin{tabularx}{0.48\textwidth}{|l|l|XX|XX|X|}
		\hline
		Method             &  Sup.    & $\varepsilon^{\rm{acc}}_{\rm{DBE}}$$\downarrow$  &  $\varepsilon^{\rm{comp}}_{\rm{DBE}}$$\downarrow$ & $\varepsilon^{\rm{plan}}_{\rm{PE}}$$\downarrow$ & $\varepsilon^{\rm{orie}}_{\rm{PE}}$$\downarrow$ & AbsRel$\downarrow$  \\ \hline
		Wei Yin, et al.\cite{yin2021learning} &   $\surd$    &  1.90         & 5.73  & 2.0  & 7.41        & 0.079 \\ \hline
		
		Monodepth2\cite{godard2019digging} & $\times$ &  \textbf{4.455} & 68.127         & 12.160           & 30.924  & \textbf{0.220}        \\ 
		$\mathrm{P}^{2}\mathrm{Net}$\cite{yu2020p}  & $\times$ &  4.922 & 67.833         & 10.823          & 28.783         & 0.241 \\
		$\mathrm{P}^{2}\mathrm{Net}$-finetune  & $\times$ &  4.628 & \textbf{49.926}         & 10.322           & 28.750         & 0.232 \\
		Ours        & $\times$  &  4.611         & 67.828  & \textbf{9.669}  & \textbf{27.215}        & 0.227 \\ \hline
	\end{tabularx}
	\newline
	\caption{IBims-1 results with the trained model on NYUv2.}
	\label{tab:IBIMS}
\end{table}

%%%%%%%%% SUPPLYMENT END

\end{document}